\documentclass[conference]{IEEEtran}
\IEEEoverridecommandlockouts
\usepackage{cite}
\usepackage{amsmath,amssymb,amsfonts}
\usepackage{algorithm}

\usepackage{algpseudocode}
\usepackage{graphicx}
\usepackage{makecell}
\usepackage{multicol}

\usepackage{subfigure}

\usepackage{textcomp}
\usepackage{xcolor}

\def\BibTeX{{\rm B\kern-.05em{\sc i\kern-.025em b}\kern-.08em
    T\kern-.1667em\lower.7ex\hbox{E}\kern-.125emX}}
\begin{document}

\title{Policy Diversity for Cooperative Agents}

\author{\IEEEauthorblockN{1\textsuperscript{st} Mingxi Tan}
\IEEEauthorblockA{\textit{Ubisoft La Forge} \\
\textit{Ubisoft}\\
Chengdu, China \\
ming-xi.tan@ubisoft.com}
\and
\IEEEauthorblockN{2\textsuperscript{nd} Andong Tian}
\IEEEauthorblockA{\textit{Ubisoft La Forge} \\
\textit{Ubisoft}\\
Chengdu, China  \\
an-dong.tian@ubisoft.com}
\and
\IEEEauthorblockN{3\textsuperscript{nd} Ludovic Denoyer}
\IEEEauthorblockA{\textit{Ubisoft La Forge} \\
\textit{Ubisoft}\\
Bordeaux, France \\
ludovic.denoyer@ubisoft.com}
}

\IEEEoverridecommandlockouts
\IEEEpubid{\makebox[\columnwidth]{paper accepted by IEEE CoG2023 \hfill} 
\hspace{\columnsep}\makebox[\columnwidth]{ }}

\maketitle

\IEEEpubidadjcol

\begin{abstract}
Standard cooperative multi-agent reinforcement learning (MARL) methods aim to find the optimal team cooperative policy to complete a task. However there may exist multiple different ways of cooperating, which usually are very needed by domain experts. Therefore, identifying a set of significantly different policies can alleviate the task complexity for them. Unfortunately, there is a general lack of effective policy diversity approaches specifically designed for the multi-agent domain. 
In this work, we propose a method called Moment-Matching Policy Diversity to alleviate this problem. This method can generate different team policies to varying degrees by formalizing the difference between team policies as the difference in actions of selected agents in different policies. 
Theoretically, we show that our method is a simple way to implement a constrained optimization problem that regularizes the difference between two trajectory distributions by using the maximum mean discrepancy. The effectiveness of our approach is demonstrated on a challenging team-based shooter.         
\end{abstract}


\section{\textbf{Introduction}}
In reel-time strategy (RTS) team-based video games, game designers often need a team of bots to challenge human-players with obviously different cooperative strategies to make the game more interesting. Traditional approaches to multi-agent collaboration diversification are usually based on complex scoring systems working with behavior trees to influence the whereabouts and behavior of each agent.
In order to design an effective scoring system, game designers need to consider complex information about bots, human-players and game events such as locations of bots and human players, weapons used, skills used, health status, different play styles, etc., which is very challenging. And it is even more challenging for them to implement different cooperative strategies through the scoring system.

Recently, Multi-Agent Reinforcement Learning (MARL) is widely used in RTS team-based video games to automatically train multiple agents for efficient cooperation and achieved prominent performance \cite{openai-2018,
VDN,QTRAN,QMIX,QPLEX,CD-PLEX}. But emerging cooperative MARL approaches such as value-based approaches \cite{QMIX,QTRAN,QPLEX,CD-PLEX} and policy-based approaches \cite{MADDPG,COMA,TRUST}, mainly aim at finding the optimal cooperative policies rather than diversify them, although this is very much needed for game designers.

We argue that the difference in the actions of the same agent in the same scenario capture well the kinds of different policies we are seeking. Therefore, we propose an algorithm denoted as Moment-Matching Policy Diversity (MMPD) to diversify the cooperative policies by changing the actions of a set of agents. We motivate MMPD theoretically by showing that it is an easy way to implement a specific constrained optimization problem that takes the Maximum Mean Discrepancy (MMD) as a constraint on the optimization objective such that the trajectory of the new policy is significantly different from that of known policies.    

The main contribution of this paper is to propose a novel policy diversification method for MARL that is different from existing policy diversification methods in the single-agent setting. To evaluate our approach, we create a challenging mini team shooter where the complex team cooperation is indispensable. The experimental results show that our approach is effective.

\section{\textbf{Related Works}}

\textbf{Policy Diversification} has been widely studied in the single-agent field. CCPT \cite{CCPT} diversifies the policies by encouraging the agent to visit less visited states. Its effectiveness has been proven in bug finding for video games. DIPG \cite{MMD} generates different policies by adding Maximum Mean Discrepancy (MMD) as a weighted regularization term to the policy-based objective and maximizes the difference between the new policy and the set of known policies while completing the same trajectory. To generate multiple significantly different policies, it requires a long trial-and-error loop to find the appropriate weights for MMD term. RSAC \cite{RSAC} applies Kullback-Leibler divergence to measure the difference between the action distributions from the new policy and the known policy, and integrates it into Soft Actor-Critic through constrained optimization framework. It can generate different policies by partially imitating a given behavior to varying degrees. The experiments on Rabidds\footnote{https://en.wikipedia.org/wiki/Raving\_Rabbids} video games have demonstrated its effectiveness. DIAYN \cite{diayn} diversifies the policies by encouraging the agent to complete the task with different skills. It uses a skill discriminator to predict the next skill of new policy and compare it to that of known policies. Their difference works as an auxiliary reward to encourage the new policy to choose the 
action that yields different skills.

However, changing the skill of agent doesn't guarantee the different way of cooperating in MARL. For example, in a team-based shooter, teams of bots can employ the same cooperative strategy whether they all use pistols or rifles. 

Different from all the methods above, our method generates different policies by making a set of agents behave differently from known policies, which is unique to MARL environment.

\section{\textbf{Background}}
A Markov game can be defined by a tuple $(K,\textbf{S},\textbf{A},P,r,\gamma)$, where $K\equiv \{1,...,k\}$ is a finite set of agents, $\textbf{s}\in \textbf{S}$ is a finite global states, 
$\textbf{A}$ is the joint-action space,
$r(\textbf{s},\textbf{a}):\textbf{S} \times \textbf{A} \rightarrow \mathbb{R}$ is the global reward function, $P:\textbf{S}\times\textbf{A}\times\textbf{S}\rightarrow [0,1]$ is the transition probability function, $\textbf{s}'\sim P(.|\textbf{s},\textbf{a})$ is a transition to the next global state and $\gamma \in [0,1)$ is the discount factor.

In this work, we consider a fully-cooperative multi-agent task where each agent has its own observation $s$ and constructs its own policy $\pi(\cdot|s)$.
All agents share the same reward function. At each time step $t$, every agent $k\in K$ chooses an action $a_t^k\in A^k$ from its individual policy
$\pi^k(a^k_t|s_t^k)$
forming a joint-action $\boldsymbol{a_t}\equiv \{a_t^1,...,a_t^k\}\in \textbf{A}$, receives the same global reward $r_t(\boldsymbol{s_t},\boldsymbol{a_t})$ and moves to next state $\boldsymbol{s_{t+1}}\equiv\{s_{t+1}^1,...,s_{t+1}^k\}$ with probability $P$. 
The objective is to learn a set of $N$ significantly different joint-policies $\Pi=\left\{\boldsymbol{\pi_i}(.|\boldsymbol{s})\equiv \{\pi_i^1,...,\pi_i^k\}\right\}_{i=1}^{N}$ 
to maximize the expected discounted returns:
\begin{equation} \label{eq3}
\begin{aligned}
& \mathop{max\mathcal{J}}\limits_{\boldsymbol{\pi_i}\in\Pi}(\boldsymbol{\pi_i})=\mathbb{E}\left[\sum\limits_{t=0}^\infty \gamma^t r_t\right].
\end{aligned}
\end{equation}

\section{\textbf{Our Approach}}
In the MARL setting, we propose a method for generating diverse cooperative joint-policies by making a set of agents behave differently from a set of known policies. In this section we will first introduce how to generate a single different joint-policy with different agent sets and different known policy sets, then how to iteratively generate multiple different joint-policies.
\subsection{\textbf{Single Joint-Policy Generation}}
To generate a new joint-policy different from the known policy, we propose an approach called Moment-Matching Action Diversity shown in \textbf{Algorithm \ref{alg:a1}}. 

\begin{figure}[b]
\centering
\subfigure[$L=\{agent_1\}$]{\includegraphics[width=4.0cm]{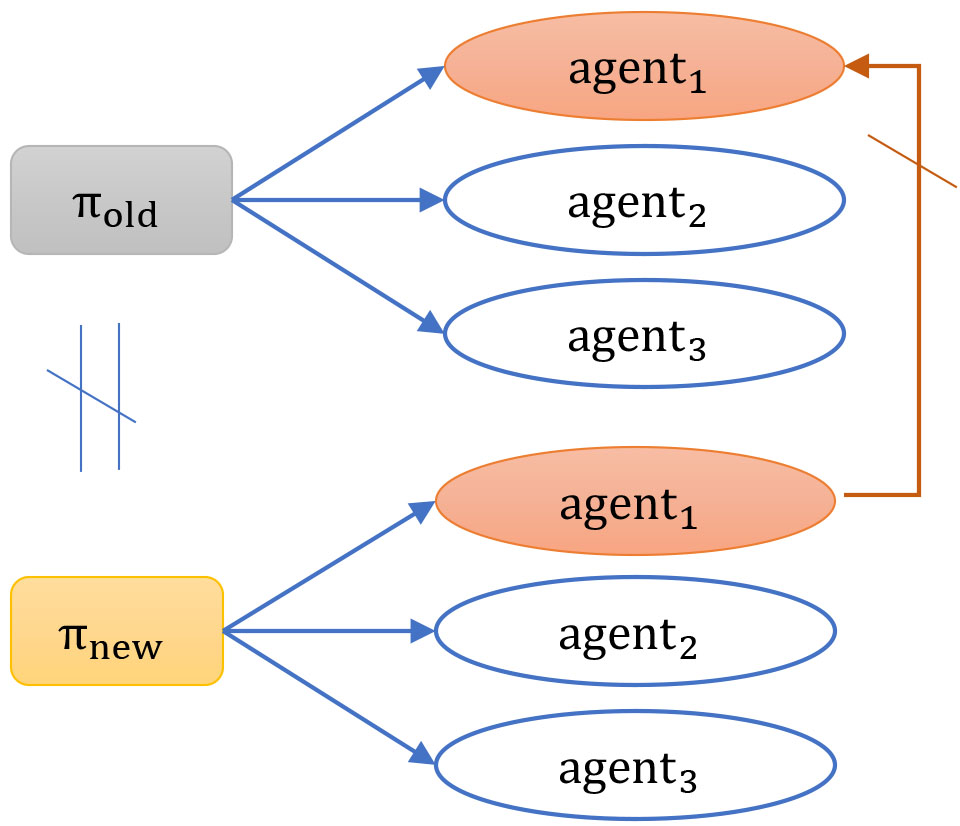}}
\subfigure[$L=\{agent_1, agent_2\}$]{\includegraphics[width=4.0cm]{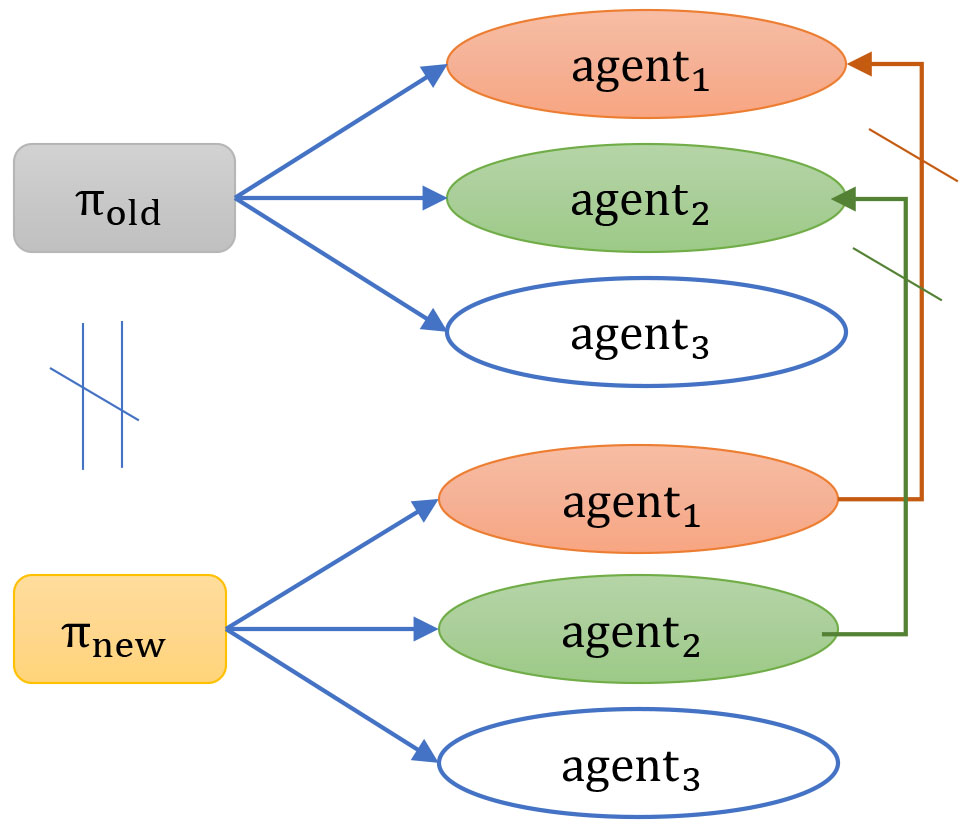}}
\caption{Generate different joint-policies by choosing different agents. (a) Generate a new policy $\pi_{new}$ by making the actions of agent$_1$ different from that of the known policy $\pi_{old}$. (b) Generate a new policy $\pi_{new}$ by making the actions of agent$_1$ and agent$_2$ different from that of the known policy $\pi_{old}$.}
\label{fig_L}
\end{figure}

It first collects $N$ transition samples by applying the known joint-policy $\pi_{kwn}$ to accomplish the task and places them in buffer $\mathcal{D}$ as $\mathcal{D}=\left\{\left\{\overline{\boldsymbol{s}}_t,\{\overline{a}^{k}_t\}_{k=1}^K, r_t, \overline{\boldsymbol{s}}_{t+1}\right\}_{t=1}^N \sim \pi_{kwn}\right\}$, shown in the input. Then it selects $L$ agents and calculates their actions through the new policy $\pi_{\theta}$ with the known state $\overline{\boldsymbol{s}}_t$ at each moment $t$ as $\{a_t^l\}_{l=1}^L=\left\{\mathop{argmax}\limits_{a^l\in A} \pi_{\theta}
(\overline{\boldsymbol{s}}_t,a^l)\right\}_{l=1}^L$, 
shown from line 3 to line 5. Finally, it makes each agent's action different from the known policy's action as $\{a_t^{l}\}_{l=1}^L\neq\{\overline{a}_t^l\}_{l=1}^L$, 
which will make the new policy significantly different from the known policy, shown from line 6 to line 7. Different choice of agent set $L$ can generate different policies, which is demonstrated in Figure \ref{fig_L}.

\begin{figure}
\centering
\footnotesize
\includegraphics[width=7.0cm]{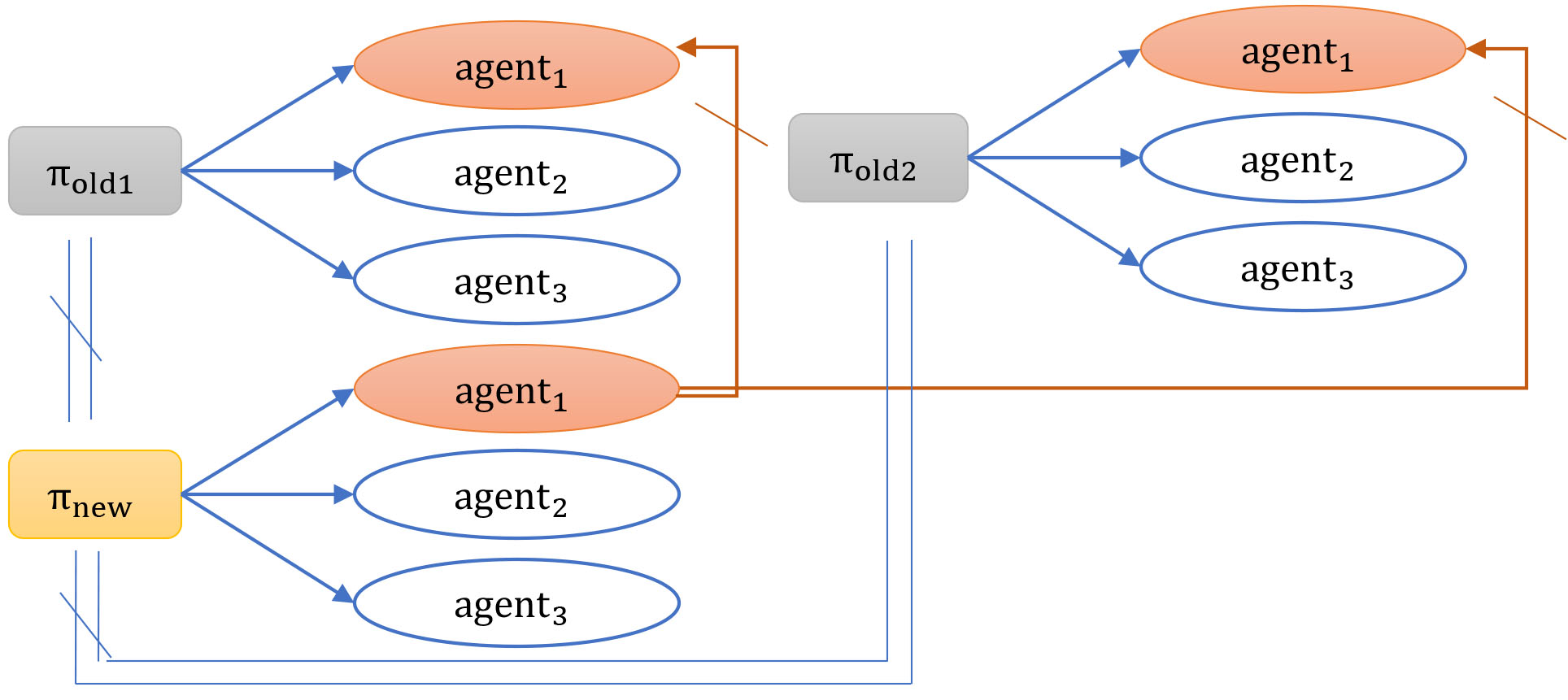}
\\{$L=\{agent_1\}$, $\Pi_{set}=\{\pi_{old_1}, \pi_{old_2}\}$}%
\caption{Generate a different joint-policy by placing more known policies in the known policy set $\Pi_{set}$. We put two known policies $\pi_{old_1}$ and $\pi_{old_2}$ in $\Pi_{set}$ to make the agent$_1$'s actions from the new policy $\pi_{new}$ different from that of $\pi_{old_1}$ and $\pi_{old_2}$.}
\label{fig_M}
\end{figure}

We can also generate a different joint-policy by placing more known policies in the known policy set $\Pi_{set}$ so that the actions of agent in $L$ will be different from the actions of each policy in $\Pi_{set}$, which is demonstrated in Figure \ref{fig_M}.

\begin{algorithm}[H]
\caption{Moment-Matching Action Diversity}\label{alg:a1}
\begin{algorithmic}[1]
\renewcommand{\algorithmicrequire}{\textbf{Input:}}
\renewcommand{\algorithmicensure}{\textbf{Output:}}

{\footnotesize

\Require Policy set with $M$ known joint-policies $\Pi_{set}$ = $\{\pi_{kwn}^m\}_{m=1}^M$, transition buffer of $M$ known policies $\mathcal{D}$=$\left\{\left\{\boldsymbol{s}^m_t,\{a^{mk}_t\}_{k=1}^K, r_t, \boldsymbol{s}^m_{t+1}\right\}_{t=1}^N \sim \pi_{kwn}^m\right\}^M_{m=1}$, replay buffer for new policy $\mathcal{D}_{new}$, selected $L$ agents from the total $K$ agents, task reward $r$ and penalty reward $\overline{r}$
\State {Initialize all the trainable parameters $\boldsymbol{\theta}$}
\Repeat
\State $\{\overline{\boldsymbol{s}}_t,\{\overline{a}^{k}_t\}_{k=1}^K, r_t, \overline{\boldsymbol{s}}_{t+1}\}_{t=1}^N \sim \mathcal{D}$ \Comment{Samples of a known policy}
\For {t=1,..,N}
\State $\{a_t^l\}_{l=1}^L=\left\{\mathop{argmax}\limits_{a^l\in A} \pi_{\theta}
(\overline{\boldsymbol{s}}_t,a^l)\right\}_{l=1}^L$ \Comment{Actions of $\pi_{\theta}$}
\If{$\{\overline{a}_t^{l}\}_{l=1}^L\neq\{a_t^l\}_{l=1}^L$ for each $l$ }{ $pass$}
\Else{ $r_t= r_t - \overline{r}$ and $\mathcal{D}_{new} \cup (\overline{\boldsymbol{s}}_t,\{\overline{a}^{k}_t\}_{k=1}^K, r_t, \overline{\boldsymbol{s}}_{t+1})$}
\EndIf
\EndFor
\State {Update $\boldsymbol{\theta}$ with SAC and $D_{new}$}

\Until {Convergence}
\Ensure $\boldsymbol{\theta}$
}
\end{algorithmic}
\end{algorithm}

\subsection{\textbf{Moment-Matching Policy Diversity (MMPD)}}

Our goal is to find $N$ significant different joint-policies. Therefore, we propose an approach noted as Moment-Matching Policy Diversity (MMPD) which combines Algorithm \ref{alg:a1}. It first trains the optimal joint-policy by setting $\Pi_{set}=\emptyset$ and $L=\emptyset$, then iteratively choose trained policies and agents and places them into $\Pi_{set}$ and $L$ respectively to train other $N-1$ joint-policies. The algorithm is listed in Algorithm \ref{alg:a2}
\begin{algorithm}[H]
\caption{Moment-Matching Policy Diversity (MMPD)}\label{alg:a2}
\begin{algorithmic}[1]
\renewcommand{\algorithmicrequire}{\textbf{Input:}}
\renewcommand{\algorithmicensure}{\textbf{Output:}}
{\footnotesize

\Require Policy set of total known joint-policies $\Pi_{all}$, subset of known joint-policies $\Pi_{set}$, transition buffer of known policies $\mathcal{D}$, replay buffer for new policies $\mathcal{D}_{new}$, agent index $K$, penalty reward $\overline{r}$ and number of new policies $N$
\State {$\Pi_{all} \leftarrow \emptyset, \Pi_{set} \leftarrow \emptyset$}
\For {n=1,...$N$}
\State $L\sim K$ \Comment{Select $L$ agents}
\State $\Pi_{set} \sim \Pi_{all}$ \Comment{Select known policies}
\State {$\pi_{\theta}$ $\leftarrow$ Algorithm \ref{alg:a1}($\Pi_{set}$, $\mathcal{D}$,$\mathcal{D}_{new}$,$L$, $\overline{r}$)} \Comment{Find a new joint-policy}
\State {$\Pi_{all}\leftarrow \Pi_{all}\cup \pi_{\theta}$}
\EndFor
\Ensure $\Pi_{all}$
}
\end{algorithmic}
\end{algorithm}

\section{\textbf{Connection to Maximum Mean Discrepancy}}
Maximum Mean Discrepancy (MMD) is a measure of the difference between 2 distributions $p$ and $q$, which is defined as:
\begin{equation} \label{eq11}
\begin{aligned}
& MMD(p,q,H)=\mathop{sup}\limits_{h\in H:\Vert h\Vert_H \leq 1}\left(E_{x\sim p}[h(x)]-E_{y\sim q}[h(y)]\right)\\
&=\left(E[k(x,x')]-2E[k(x,y)]+E[k(y,y')]\right)^{\frac{1}{2}},
\end{aligned}
\end{equation}
where $k$ is a kernel, $h$ is a test function from a function space $H$ which is a unit-ball in a Reproducing Kernel Hilbert Space (RKHS). It can be used to measure the difference between 2 trajectory distributions produced by 2 different policies.

We first let a set of $N$ state samples from different moment $t$ be defined as $\{s_t\}^N_{t=1}$. Then the moment-matching state-action samples of a total of $K$ agents from policy $p$ and $q$ can be found separately as: 
\begin{equation} \label{eq12}
\begin{aligned}
&\tau^p=\left\{s_t,\{a^k_t\sim p(.|s_t)\}_{k=1}^K\right\}_{t=1}^N, \\
&\tau^q=\left\{s_t,\{a^k_t\sim q(.|s_t)\}_{k=1}^K\right\}_{t=1}^N.
\end{aligned}
\end{equation}
where $\tau^p$ and $\tau^q$ have the same states. 
Then, we can continue define a kernel in Eq.\ref{eq11} (such as Gaussian kernel) over moment-matching samples $(\tau^p, \tau^q)$ as:
\begin{equation} \label{eq13}
\begin{aligned}
k(\tau^p, \tau^p)=K(\phi(g(\tau^p)), \phi(g(\tau^q))),
\end{aligned}
\end{equation}
where $g$ is a function to map every action-state pair in $\tau$ to a value, since $\tau^p$ and $\tau^q$ have the same states, $g$ can be simplified to map every action to a value. $\phi$ simply stacks the state-action pairs into a single vector. Let policy $p$ be the reference policy, then every action chosen from $p$ can get a constant value 1 from $g$ as:
\begin{equation} \label{eq14}
g(a^{{ref}_k}_t)=1 \text{ If $a^{{ref}_k}_t \sim p$}
\end{equation}
For the action chosen from another policy such as $a^k_t\sim q$, it can obtain a value from $g$ as
\begin{equation} \label{eq15}
g(a^k_t)=
\begin{cases}
1&\text{if $a^{{ref}_k}_t = a^k_t $}\\
0&\text{if $a^{{ref}_k}_t \ne a^k_t $}
\end{cases}
\end{equation}

Finally, we can calculate the MMD between $\tau^p$ and $\tau^q$ through  MMD$(\tau^p, \tau^q;K)$ with Eq.\ref{eq11}. If $\{a^k_t\sim p(.|s_t)\}_{k=1}^K=\{a^k_t\sim q(.|s_t)\}_{k=1}^K$ for every agent k at each moment $t$, $\tau^p$ will be the same as $\tau^q$
leading the minimum of MMD. If $\{a^k_t\sim p(.|s_t)\}_{k=1}^K\ne\{a^k_t\sim q(.|s_t)\}_{k=1}^K$ for every agent k at each moment $t$, MMD will reach the maximum. 

Algorithm \ref{alg:a1} attempts to find a significantly different policy by making every agent $l$ ($l \in L$) behaves differently from every known policy $m$ ($m \in M$) for every moment $t$, as shown below:
\begin{equation} \label{eq16}
\begin{aligned}
&\left\{a^l_t\sim argmax(\pi_{\theta}\left(.,\textbf{s}_t)\right)\right\}_{l=1}^L\\\
&\ne\left\{\left\{a^l_t\sim argmax(\pi^m_{kwn}(.,\textbf{s}_t))\right\}_{l=1}^L\right\}_{m=1}^M.
\end{aligned}
\end{equation}
So algorithm \ref{alg:a1} is essentially equivalent to changing the original objective in Eq.\ref{eq3} to an MMD-based constraint objective $\mathcal{J}_{MMD}(\pi)$ shown as: 
\begin{equation} \label{eq17}
\begin{aligned}
&\mathop{max}\limits_{\pi}\mathcal{J}_{MMD}(\pi)=\mathop{max}\limits_{\pi}\mathcal{J}(\pi)\\
&s.t. \left\{MMD(\tau^{\pi_{\theta}}, \tau^{\pi_{kwn}^m};L)\ge \overline{MMD}_m\right\}_{m=1}^M,
\end{aligned}
\end{equation}
where $\overline{MMD}_m$ is a predefined value that guarantees that the new joint-policy is significantly different from the $M$ known polices. To solve Eq.\ref{eq17}, we need to first formulate it as a dual problem by adding Lagrangian multipliers and then apply the Min-Max mechanism to solve both the policy and the Lagrangian multipliers \cite{RSAC}. Additionally, it usually requires a long trial-and-error loop to find the right $\overline{MMD}_m$. However Algorithm \ref{alg:a1} is straightforward in comparison. 

\section{\textbf{Experimental Setup}}
To evaluate the proposed method, we build a challenging mini team-based shooter, where 2 white cubes fight against 1 red cube which can only turn in place and is controlled by Behavior-Trees to shoot the nearest white cube. 

Each white cube has 2 life points and two skills: using gun and using bomb. To use the gun, the white cube can shoot the red cube from 20 meters away for every 2 seconds and deal 1 health point per shot. To use the bomb, the white cube can attack the red cube from 15 meters away for every 3 seconds and deal 2 health points per attack. 

The red cube has 8 life points, which means the white cubes can defeat the red cube with 8 shots from the gun or 4 attacks with the bomb. The red cube can aim at the nearest white cube from 100 meters away and shoot from 20 meters away. Each of its shots reduces the health of the white cube by 1 point. So the red cube is much stronger than the white cube that one white cube can never beat it. Therefore, 2 white cubes need to cooperate to win the game. 

\begin{figure}[H]
\centering
\subfigure[Trajectories of using gun (yellow) vs using bomb (red)]{\includegraphics[width=4.3cm]{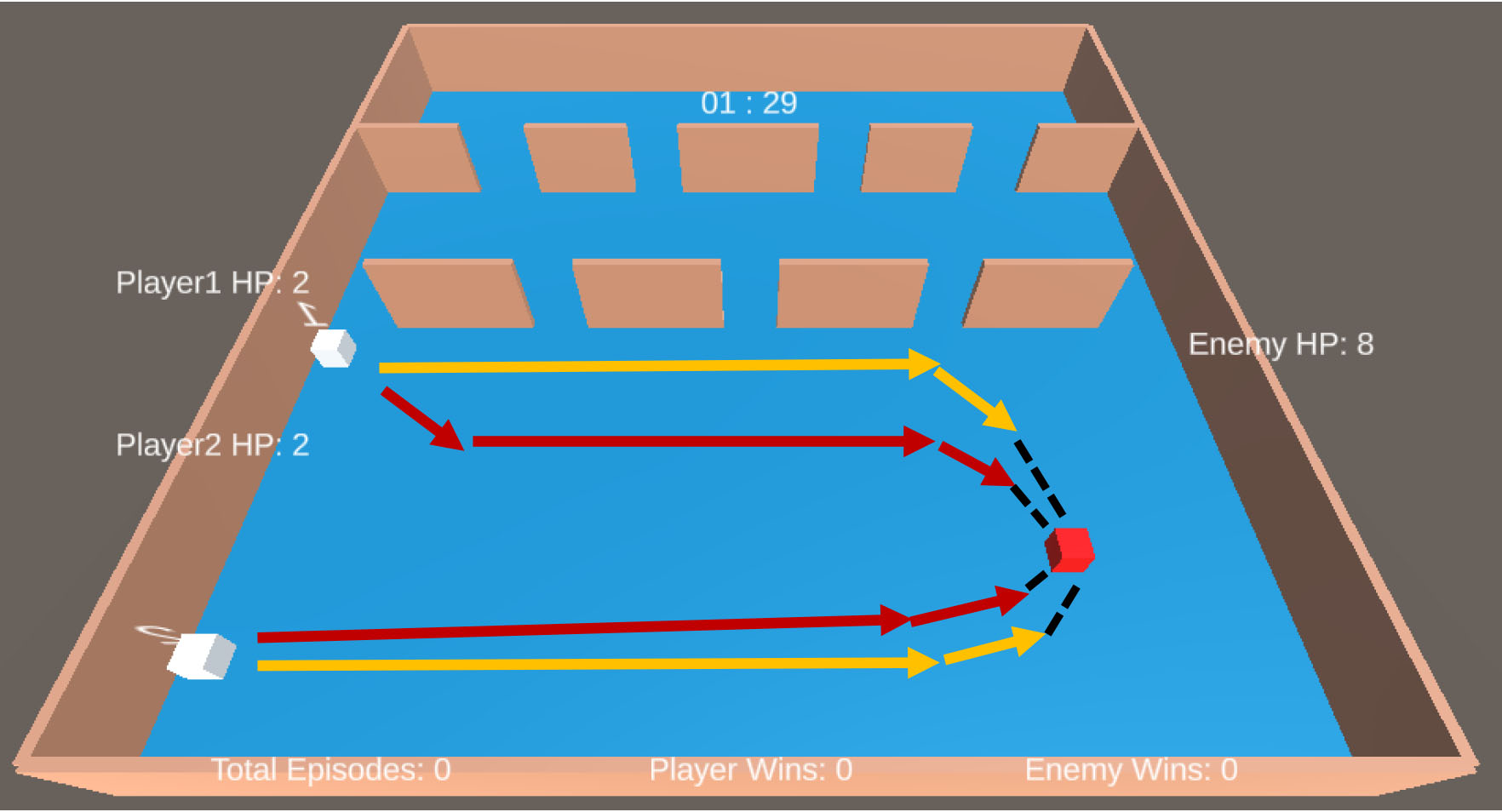}}
\subfigure[Trajectories of $L_1$ (yellow) vs $L_2$ (red)]{\includegraphics[width=4.3cm]{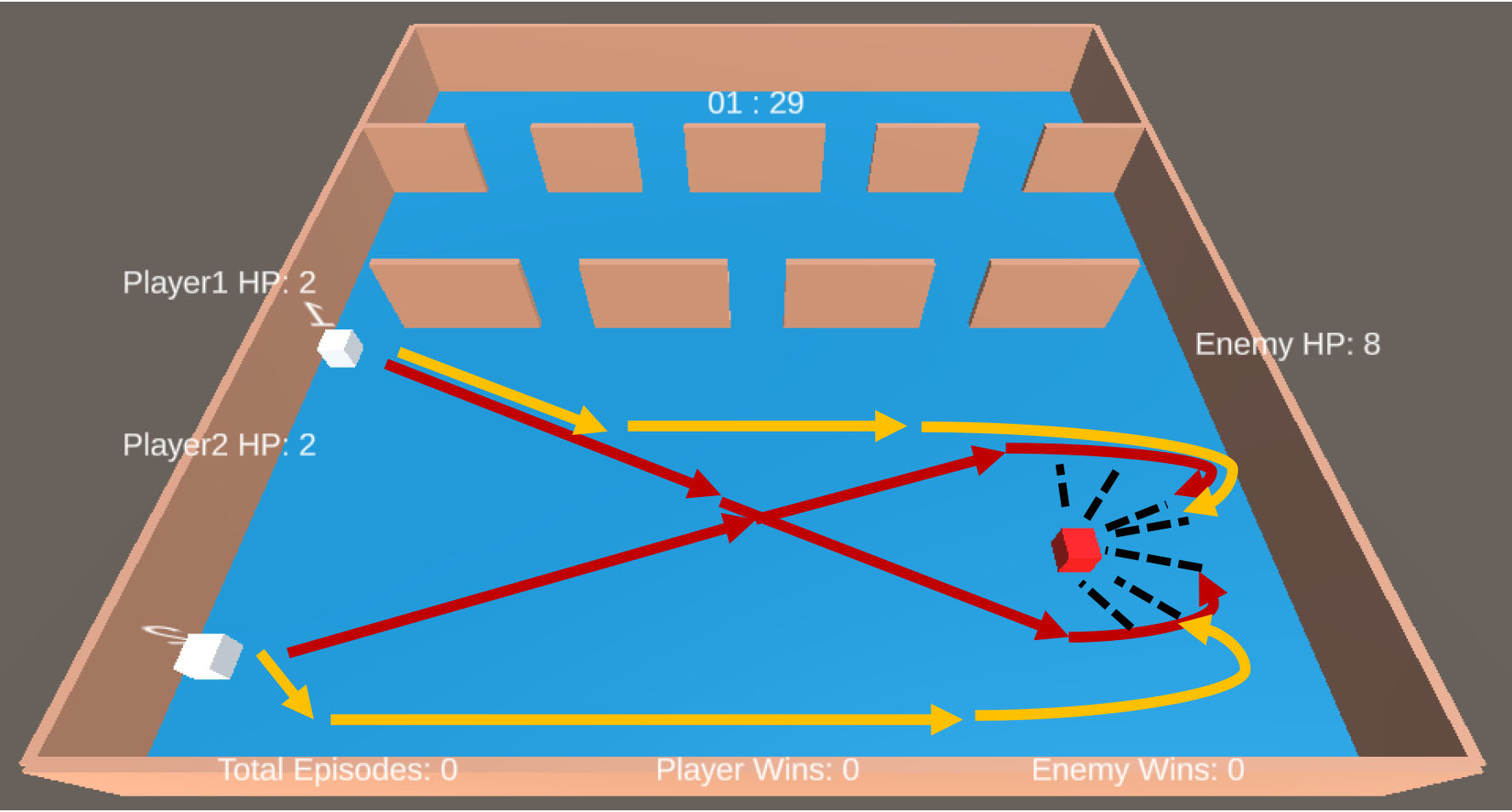}}
\caption{Trajectories of $agent_1$ and $agent_2$ obtained from different approaches. a) Gun vs Bomb. b) $L_1=\{agent_1\}$ vs $L_2=\{agent_1, agent_2\}$. In sub-figure (a), we can see that the agents' trajectories are similar when the agents use different skills. However in sub-figure (b), agents' trajectories are much more different by using our approach. The yellow and red lines represent the trajectories of different methods. The black dotted line represents the attack action of agent.}
\label{fig2}
\end{figure}

We first train two joint-policies by making the white cube only use gun or bomb. Then we train two joint-policies by using different agent sets: $ L_1=\{agent_1\}$ and $L_2=\{agent_1, agent_2\}$. Finally, we  train other two different policies by using different known policy sets: $\Pi_{set_1}=\{\pi_1\}$ and $\Pi_{set_2}=\{\pi_1, \pi_2\}$. Every model is trained with five random seeds. We use Fréchet Distance \cite{Frechet-Distance} to compare the difference in trajectories of the same agent between different policies, the average value of which is shown in Table \ref{tab1} and an example of their trajectory is shown in Fig \ref{fig2}.

Based on the results, we can make several important observations: 
\begin{itemize}
    \item [1.]
    In MARL, the team policies could be similar when the agents apply different skills (second row in table \ref{tab1} and sub-figure (a) in Fig \ref{fig2});
    \item [2.]
    Our approach (MMPD) can generate significantly different team policies by using different agent sets and different known policy sets (Third and forth row in Table \ref{tab1} and sub-figure (b) in Fig \ref{fig2}). 
\end{itemize}


\begin{center}
\begin{table}[H]
        \caption{Trajectory Comparison with Fr\'echet Distance \\
        (the higher the better)}
        \centering
        \resizebox{.5 \textwidth}{!}
        {
        \begin{tabular}{|c|c|c|}
            \hline
              \makecell{ } & \textbf{For} $\boldsymbol{agent_1}$ & \textbf{For} $\boldsymbol{agent_2}$ \\
            \hline
            $Skill_{Gun}$ vs $Skill_{Bomb}$ $^{\mathrm{a}}$ & 1.28 & 0.80  \\ 
            \hline
            $MMPD_{\Pi_{set_1}}$ vs $MMPD_{\Pi_{set_2}}$ $^{\mathrm{b}}$ & 2.80 & 2.40\\ 
            \hline
            $MMPD_{L_1}$ vs $MMPD_{L_2}$ $^{\mathrm{c}}$ & 3.20 & 4.40\\ 
            \hline
            \multicolumn{3}{l}{$^{\mathrm{a}}$The first row shows the difference of Fr\'echet Distance in trajectories} \\
            \multicolumn{3}{l}{{ }{ }using different skills.} \\
            \multicolumn{3}{l}{$^{\mathrm{b}}$The second row shows the difference of Fr\'echet Distance in  trajectories} \\
            \multicolumn{3}{l}{{ }{ }using different policy sets.} \\
            \multicolumn{3}{l}{$^{\mathrm{c}}$The third row shows the difference of Fr\'echet Distance in trajectories} \\
            \multicolumn{3}{l}{{ }{ }using different agent sets.}
        \end{tabular}
        }
        \label{tab1}
\end{table}
\end{center}

\section{\textbf{Conclusion}}
In this paper, we propose an approach of identifying a set of significantly different multi-agent cooperative policies by making the selected agents behave differently from that of known policies, which is unique for MARL setting. We implement our approach with SAC and apply it to a challenging team-based shooter where the team cooperation is essential. Experimental results show that our approach is effective in diversifying cooperative policies.


\vspace{12pt}

\end{document}